# Machine Learning and AI Applied to fNIRS Data Reveals Novel Brain Activity Biomarkers in Stable Subclinical Multiple Sclerosis

Sadman Saumik Islam, Bruna Dalcin Baldasso, Davide Cattaneo, Xianta Jiang and Michelle Ploughman

***Abstract*—People with Multiple Sclerosis (MS) complain of problems with hand dexterity and cognitive fatigue. However, in many cases, impairments are subtle and difficult to detect. Functional near-infrared spectroscopy (fNIRS) is a non-invasive neuroimaging technique that measures brain hemodynamic responses during cognitive or motor tasks. We aimed to detect brain activity biomarkers that could explain subjective reports of cognitive fatigue while completing dexterous tasks and provide targets for future brain stimulation treatments. We recruited 15 people with MS who did not have a hand (Nine Hole Peg Test [NHPT]), mobility (fast walking speed), or cognitive (Montreal Cognitive Assessment) impairment, and 12 age- and sex-matched controls. Participants completed two types of hand dexterity tasks with their dominant hand, single task (NHPT) and dual task (NHPT while holding a ball between the fifth finger and hypothenar eminence of the same hand). We analyzed fNIRS data (oxygenated and deoxygenated hemoglobin levels) using a machine learning framework to classify MS patients from controls based on their brain activation patterns in bilateral prefrontal and sensorimotor cortices. The K-Nearest Neighbor classifier achieved an accuracy of 75.0% for single manual dexterity tasks and 66.7% for the more complex dual manual dexterity tasks. Using explainable AI, we found that the most important brain regions contributing to the machine learning model were the supramarginal/angular gyri and the precentral gyrus (sensory integration and motor regions) of the ipsilateral hemisphere, with suppressed activity and slower neurovascular response in the MS group. During both tasks, deoxygenated hemoglobin levels were better predictors than the conventional measure of oxygenated hemoglobin. This nonconventional method of fNIRS data analysis revealed novel brain activity biomarkers that can help develop personalized brain stimulation targets.**

The work was partially supported by Natural Sciences and Engineering Research Council of Canada (NSERC, Grant No DGECR-2020-00296). The work of Michelle Ploughman was partially supported by Canada Research Chairs (950-232532), Canadian Institutes for Health Research (173526), Newfoundland and Labrador Research and Development Corporation (5404.1699.104), Canada Foundation for Innovation (33621).

Sadman Saumik Islam and Xianta Jiang are with the Department of Computer Science, Memorial University of Newfoundland, St. John's, NL A1B 3X5, Canada.

Bruna Dalcin Baldasso and Michelle Ploughman are with the Division of Biomedical Science, Memorial University of Newfoundland, St. John's, NL A1B 3X5, Canada.

Davide Cattaneo is with the Department of Pathophysiology and Transplantation, Fondazione Don Carlo Gnocchi, Milan, Italy

## I. INTRODUCTION

MANY people with Multiple sclerosis (MS) complain of problems about hand dexterity and cognitive fatigue. However, in many cases, impairments are subtle and difficult to detect [1]. Early diagnosis and advances in high-efficacy disease-modifying treatments have led to fewer relapses and better outcomes among people with MS [2]; however, gradual accumulation of impairment can occur in the absence of relapses [3]. Novel tools and methods are needed to index impairments, such as those involving fatigue and hand dexterity, not easily seen by a clinician or outside observer [4] and to define their neurological underpinnings.

Coordination of multiple brain regions, including the primary motor cortex, premotor cortex, supplementary motor area, cerebellum, and basal ganglia, is required for planning, executing, and controlling hand movements and integrating sensory feedback [5]–[7]. Among healthy people, there is often bilateral cortical activation, especially in more complex tasks [7] but the hemisphere contralateral to the moving hand is usually more active than the ipsilateral side [7]. People who have a stroke tend to demonstrate ipsilateral activation [8], [9] when attempting to move the paralyzed hand, which is considered a poor prognostic sign [10]–[12]. However, MS lesions can be scattered, leading to bilateral deficits and less predictable changes in cortical activation, [13] and such network dysfunction could explain why people with MS experience cognitive fatigue when completing physical and cognitive tasks [14]. Precision rehabilitation that specifically targets disrupted brain regions holds promise. For instance, preliminary trials of brain stimulation techniques such as transcranial direct current stimulation [15], [16] may improve MS-related cognitive fatigue; however, the optimal stimulation targets are not currently known. The field of brain stimulation requires new methods to identify brain activity biomarkers that could serve as treatment targets.

Recent advances in functional brain imaging techniques have enabled researchers to investigate brain network dysfunction underlying MS and its associated symptoms [17]–[19]. However, there have been very few studies measuring cortical activation during upper limb movement in MS [20]–[22]. Some studies suggest that cortical activation is similar to healthy controls [23] while others report increased bilateral recruitment [24]. Functional Near-Infrared Spectroscopy (fNIRS) uses near-infrared light to measure changes in cortical blood oxygenation levels [25], [26]; a proxy for neural activation. Although fNIRS is very useful for detecting brain activation during motor tasks[27], [28], most studies analyzed data by taking a conventional route, which is a data-reductionist approach that only reports average oxyhemoglobin (HbO; an indicator of metabolic demand) and in some studies, also deoxyhemoglobin (HbR; an indicator of oxygen utilization) [29], [30]. By evaluating time-related data from every fNIRS channel, Machine Learning (ML) and Artificial Intelligence (AI) have the potential to counteract this reductionist approach and expose novel features, patterns, and insights that are difficult to find using traditional methods [31]. To date, no studies have employed ML and AI on fNIRS data in the field of MS.

This study employed ML algorithms to classify MS patients from controls based on fNIRS brain activation patterns during two types of hand movement (single and dual manual dexterity tasks). Importantly, participants had no overt cognitive, mobility, or hand impairment. The overarching aim was to determine whether explainable AI could be useful in identifying potential brain activity biomarkers that could inform personalized brain stimulation targets for individuals with MS.

## II. Methods

### A. Design and recruitment

According to the Declaration of Helsinki, and with the approval of the local Health Research Ethics Board (HREB#2021.005), participants with MS were recruited from a specialized MS neurology clinic and controls through poster advertisements. After providing informed written consent, eligibility was assessed through a screening interview. Additionally, health records were reviewed for individuals with MS. The inclusion criteria for MS participants were 1) diagnosed with relapsing-remitting MS by a neurologist following the McDonald criteria [32], 2) relapse-free during the previous three months, and 3) an Expanded Disability Status Scale (EDSS) score ≤ 3.0, indicating no mobility impairment. All participants were excluded if they were 1) under 18 years of age, 2) diagnosed with any other neurological conditions, 3) had untreated cardiovascular disease, which could potentially affect brain hemodynamics, 4) reported hand or arm impairment, or 5) scored lower than 26 on the Montreal Cognitive Assessment, which is considered abnormal [33], 5) left-handed, 6) reported moderate depressive symptoms, scoring more than 10 out of 21 on the depression subscale of the Hospital Anxiety and Depression Scale [34], or 7) pregnant. Controls were matched by age (±3 years) and sex. Although there was insufficient comparable research to complete a sample size calculation, we aimed to recruit at least 10 people in each group, having complete fNIRS data.

### B. Outcome measures

Nine-Hole Peg Test (NHPT) [35]. The NHPT is a validated tool for assessing upper extremity motor function, which involves allocating pegs into holes in a wooden platform and removing them individually. Participants completed the task twice with each hand, and the completion time was recorded using a stopwatch. Scores were averaged for each hand.

Edinburgh Handedness Inventory (EHI): The EHI is a validated tool for evaluating handedness (Oldfield, 1971), which requires participants to indicate their dominant hand during their day-to-day lives. Participants are then categorized into right-handed, left-handed, or mixed-handed based on their handedness.

The Montreal Cognitive Assessment (MoCA): To assess cognitive impairments, participants completed the MoCA, which tests five domains, including attention, language, memory, orientation, and visuospatial and executive function. The total score of these domains was 30 points, where scores less than 26 were considered abnormal [33].

The Hospital Anxiety and Depression Scale (HADS): The participants completed the HADS, which consists of 14 items measuring symptoms of anxiety and depression during the preceding two weeks. Only the depression subscale score was considered in this study. Each item is scored from 0 to 3 points, summing up to a maximum of 21 points for each subscale [34].

Expanded Disability Status Score (EDSS) was collected from the participant's medical charts. We used the EDSS (0-10) to measure the level of MS-related disability.

### C. fNIRS System and Optode Array

We used the continuous-wave NIRScoutX® 16 ×16 imaging system (NIRx Medical Technologies, Berlin, Germany), which contains 16 LED sources and 16 detectors to collect the fNIRS data at a sampling rate of 3.9 Hz. Eight sources and eight detectors (20 channels) were positioned over the motor cortex (Figure 1). The probe position on the cap was identified using the fNIRS Optode Location Decider (fOLD) program [36], with sensitivity ranging from 43.1% to 87.4%. Spacers were used to maintain a consistent gap of 3 cm between sources and detectors. A set of short-range detectors was attached to the optodes to measure and eliminate data regarding superficial scalp blood flow. Before testing, head circumference was measured relative to the nasion and the inion; which was used to calculate each individual's cap size. Probes were calibrated, and each channel was checked for noise and signal quality before data recording. The signal was deemed acceptable if the noise level was less than 7.5 dB and the gain was equal to or greater than 7. All data recordings were completed using the NIRStar® collection program (NIRx Medical Technologies, Berlin, Germany).

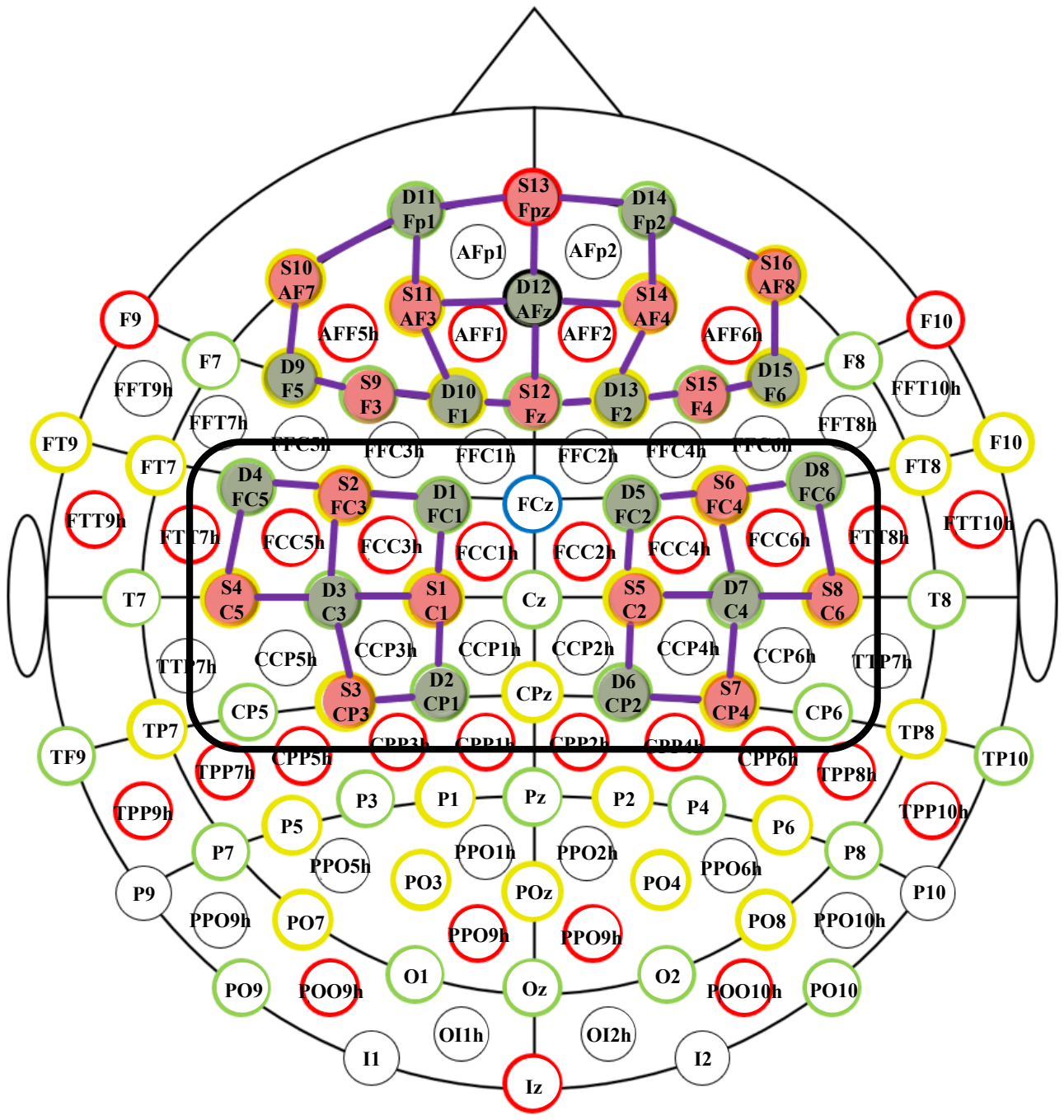

Fig. 1. Configuration of channels in the fNIRS cap covering two regions of interest (ROI) (Channels inside the black box): left motor cortex (S1-S4 and D1-D4); right motor cortex (S5-S8 and D5-D8). Each purple line represents a channel connecting a source to a detector; sources (S) are indicated by red circles, and detectors (D) are indicated by green circles.

## D. Procedure

Participants were instructed to avoid caffeine, alcohol, and smoking for 24 hours prior to the fNIRS laboratory visit. They sat comfortably facing a screen approximately 2.5m away. After receiving standardized instructions, the participants conducted a practice trial without wearing the fNIRS caps. The cap was then positioned on the head, using the inion as a reference point for accurate probe placement. To prevent excessive environmental light, ambient lights were turned off, and blackout curtains were closed.

The experiment involved completing two types of motor tasks: the single and dual manual dexterity tasks. The single task required the participant to carry out the NHPT, while during the dual task, the participant had to simultaneously hold a small ball between the fifth finger and hypothenar eminence of the same dominant hand while completing NHPT (NHPT + holding a ball). Each task lasted for 20s, followed by a 20s rest, presented in random order [25]. Instructions were displayed on the screen six seconds before the start of each task. The rest periods were indicated using auditory start and stop signals (Figure 2). The stimulus presentation was developed and randomized using the NIRStim® platform (NIRx Medical Technologies, Berlin, Germany). Each participant completed 10 task trials consisting of 5 single task trials and 5 dual task trials in a random order. For the ML models, the input window size was 20 seconds, which was the length of a trial. The evaluation metrics used were Accuracy, Precision, Recall, and F1-Score, as described in the section Evaluation criteria and cross-validation below.

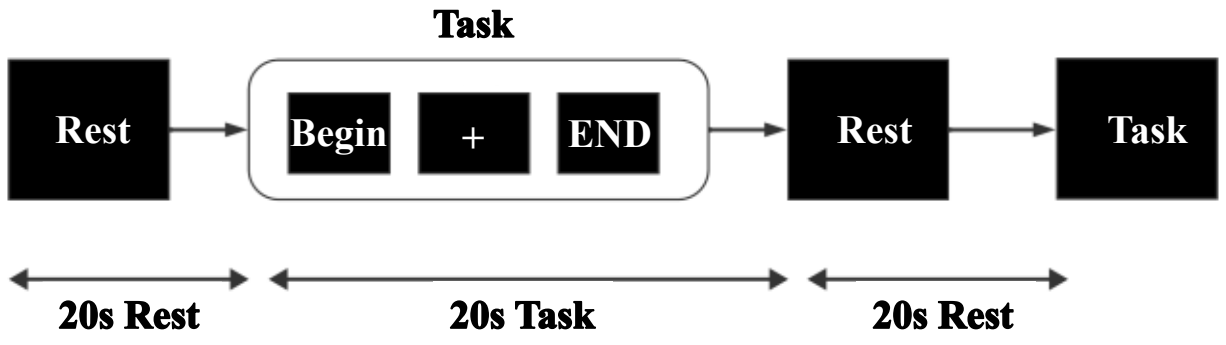

Fig. 2. Task sequence arranged in a block design. Participants executed the task for 20s, followed by 20s of rest. The task order was randomized for each participant.

## D. Data Analysis

1. Preprocessing of fNIRS data: We created a pipeline for preprocessing, which included data filtering, noise removal, and transformation of light intensity to hemoglobin concentrations. In the first step of the pipeline, short-channel data [25] were used to filter out superficial signals (e.g., scalp blood flow) from the fNIRS measurements. Then, the optical intensity was converted to HbO and HbR concentrations using the modified Beer-Lambert law [37]. Both a low-pass filter (cut-off frequency of 0.05hz) and a high-pass filter (cut-off frequency of 0.7hz) were employed to remove cardiac and respiration signals and slow drifts in the data [38]. The data preprocessing pipeline was implemented using the MNE-Python [39], [40], an open-source package based on Python, designed to explore, visualize, and analyze human neurophysiological data such as fNIRS. The version of Python was 3.11.1, and the version for the MNE-Python library was 1.6.1. Spline and Wavelet methods combined were used to correct the motion artifacts [41].
2. Machine Learning Algorithms: We implemented ML classification algorithms to differentiate the MS from control groups using the HbO and HbR concentrations recorded during the duration of task periods for each trial. The channels' data (HbO and HbR) were used as input features. We used the feature selection algorithm SelectKbest [42], [43] to help reduce irrelevant or redundant features to enhance the accuracy of the ML models while also reducing the training time and computational complexity. Four different classification algorithms were chosen based on their suitability for handling fNIRS data characteristics, including Random Forest (RF), Support Vector Machine (SVM), K-Nearest Neighbor (KNN), and Light Gradient Boosting (LightGBM).

   Random Forest (RF) [44] is an ensemble learning algorithm used for classification and regression tasks in ML. It creates many decision trees during model training, extracting each tree's mode (for classification) or average prediction (for regression). RF is very capable of handling high-dimensional and noisy data [45] such as fNIRS, as well as capturing the nonlinear and complex relationships between the features [46].

   Support Vector Machine (SVM ): SVM [47] is an ML model that can be used for classification or regression tasks. It works by finding the hyperplane that maximizes the margin between two classes of data points in a high-dimensional space.

K-Nearest Neighbors (KNN): KNN [48] is a nonparametric ML model that can be used for classification or regression tasks. It works by finding the k-closest data points to a given query point and using the labels or values of those points to predict the query point. KNN can utilize various values of k and distance metrics to achieve various levels of complexity and accuracy, and it is able to adapt to the local structure and variability of the fNIRS data.

Light Gradient Boosting (LightGBM): LightGBM [49] is a gradient-boosting library designed to be efficient and scalable. It uses histogram-based algorithms to split the data and optimize the decision trees. LightGBM is a popular choice for large-scale and sparse datasets such as fNIRS, and it can use different boosting types and learning rates to optimize performance.

3. Cross-participant validation: We used a six-fold cross-participant validation [50] to prevent the overfitting problem and generate more reliable results. In each fold validation, 4 participants (2 MS and 2 controls) were used for testing, and the rest for training the model. This single fold validation process was repeated 6 times to cover all 24 participants tested. To compare the ML models' performance for the classification task, we used Accuracy, Precision, Recall, and F1-Score. The formula for the evaluation's metrics is in the Supplementary (Figure A).
4. Explainable AI: SHAP (SHapley Additive exPlanations) [51] was used to interpret our best-performing model to identify the most relevant channels that contributed to the differentiation of MS from controls. The version of SHAP used was 0.44.0. SHAP is a game theory-based approach that explains any ML model's output using Shapley values from game theory, which is a measure of the impact of a feature on the prediction of the model. SHAP ranks the model's features by assigning an important value to each feature contributing to its prediction. It can also explain the interaction between features. To address feature interactions, SHAP measures the amount of change in prediction when two features are observed together instead of individually. In our study, the features of the models were all the channels shown in Figure 1. SHAP ranked the channels by measuring each channel's impact on the ML model's prediction performance.
5. Statistical Analysis: We used the one-way ANOVA (F-test) to exam the differences between the controls and MS group using the traditional data reductionist method, where all the channels adjacent to S1-S4 were averaged into one single region of interest for the left hemisphere of the brain, and channels adjacent to S5-S8 were averaged into one region of interest for the right hemisphere of the brain. The individual channels are mainly averaged into anatomical regions to reduce the high dimensionality of the fNIRS data.

We further used the Independent Samples T Test (two-sided) for the top four most influential channels observed from the SHAP measurement to reveal significant differences in channel importance between the groups: controls and MS. The following values are provided for each group: N was the number of observations, Mean was the average value of the measurement, and Std. deviation was the standard deviation, which measured the variation or dispersion of the values.

Levene's Test for Equality of Variances assessed whether the variances of the two groups were equal. A significant result ($P < 0.05$) suggests unequal variances.

We used a T-test for Equality of Means to compare the means of the two groups. The test was performed under two assumptions: equal variances and unequal variances. t was the t-value from the t-test. A large absolute t-value (positive or negative) indicated a significant difference between the group means, and a value close to 0 indicated no significant difference. The analysis was carried out in SPSS® (IBM version 29.0.2.0).

## III. Results

We recruited 27 participants. However, data of two participants (one MS patient and one control) were excluded due to corruption in the data. We further excluded one left-handed participant, leaving 24 participants (13 MS patients and 11 controls). Table I shows the demographic and clinical characteristics of the remaining participants. The MS patients had very mild or no visible disability, with median Expanded Disability Status Scale (EDSS) scores ranging from 0-3.  There was no statistically significant difference in age between the groups in terms of EDSS score. For the NHPT score, the MS group had a higher score on average (slower performance), though this difference was not statistically significant. In contrast, the MoCA score showed that the controls scored significantly higher than MS patients (p-value = 0.02).

TABLE I

Demographic and clinical characteristics of the participants.

| | MS (N = 13) | | Controls (N = 11) | | F | P |
|---|---|---|---|---|---|---|
| | Mean | SD | Mean | SD | | |
| Age | 43.85 | 9.57 | 44.36 | 10.56 | 0.016 | 0.9 |
| Years with MS | 15.15 | 6.9 | not applicable | not applicable | not applicable | not applicable |
| NHPT score (s) | 21.43 | 3.95 | 18.94 | 2.21 | 3.43 | 0.07 |
| MoCA score | 26.85 | 1.52 | 28.18 | 1.17 | 5.65 | 0.02* |

MS: multiple sclerosis, NHPT: Nine Hole Peg Test, MoCA: Montreal Cognitive Assessment, SD: Standard Deviation. * Significantly different p<0.05.

When comparing brain activation patterns in either the right or left hemispheres using typical statistical methods, there were no statistically significant differences between MS and controls (Supplementary Table A).

The differences in model performance for the classification task can be observed in Table II for the single task and Table III for the dual task. Table II shows that the KNN and SVM models were better for the single task than RF and LightGBM, and Table III shows that for the dual task, KNN was the best model, and SVM was the worst model. RF and LightGBM had a similar performance for the dual task. Therefore, further analysis of channel contributions was conducted based on KNN models for both tasks.

TABLE II
CLASSIFICATION MEASURES FOR SINGLE TASK.

| Classifier | Accuracy | Precision | Recall | F1-Score |
|---|---|---|---|---|
| RF | 66.70% | 70.50% | 66.70% | 62.80% |
| LightGBM | 62.50% | 55.20% | 62.50% | 56.10% |
| KNN | 75.0%* | 76.00% | 75.00% | 71.70% |
| SVM | 70.80% | 69.40% | 70.80% | 65.00% |

TABLE III
CLASSIFICATION MEASURES FOR DUAL TASK.

| Classifier | Accuracy | Precision | Recall | F1-Score |
|---|---|---|---|---|
| RF | 62.50% | 60.80% | 62.50% | 56.10% |
| LightGBM | 62.50% | 61.10% | 62.50% | 56.70% |
| KNN | 66.7%* | 75.00% | 66.70% | 65.00% |
| SVM | 54.20% | 49.70% | 54.20% | 47.20% |

As shown in Figure 3, channel importance analysis (SHAP analysis) shows similar patterns for both tasks, with the HbR of channel S7-D6 identified as the most important for both task models. Figure 3 shows the top ten channels (HbO and HbR) identified by SHAP for the KNN models for both tasks (Single and Dual). Among those ten channels, the top four have a higher magnitude of impact than the rest for both single task (Figure 3(a)) and dual task (Figure 3(b)) The top four channels for both of the tasks are situated in the ipsilateral cortex (Figure 4(a) and Figure 5(a)), and the HbR concentration of those channels has more impact on the models, except for the S6-D5 channel for the dual task where the HbO concentration of the channel (Figure 3(b)) had more impact. The SHAP values are calculated using a method from cooperative game theory called Shapley values, which fairly distribute the "payout" (model prediction) among the "players" (features) [52]. The mean SHAP value is each feature's average absolute SHAP value, showing its overall importance in the model in Figure 3.

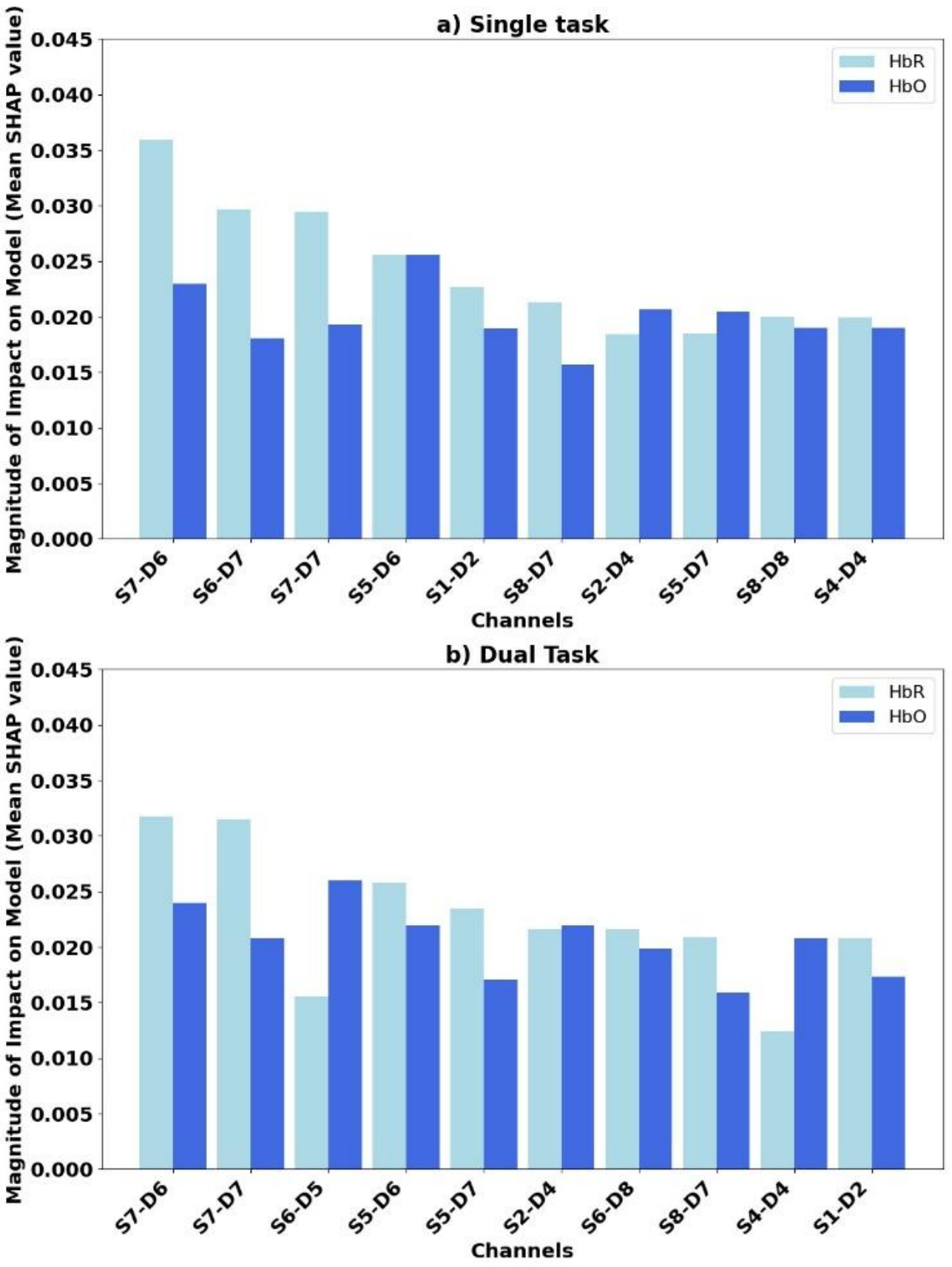

Fig. 3. The most impactful HbO and HbR channels for the K-nearest neighbor (KNN) model, identified using SHAP, under both single-task (a) and dual-task (b) conditions, respectively. Each bar represents the magnitude of the impact on the model, measured by the Mean SHAP value, for both HbO (in royal blue) and HbR (in light blue). The values on top of each bar indicate the exact SHAP values for each channel. The channels are labeled along with the x-axis, with the y-axis showing the Mean SHAP values.

Figure 4(b) illustrates the representative brain activity of one control and one MS patient during the single task, and Figure 5(b) illustrates the representative brain activity of one control and one MS patient during the dual task. For both tasks, we can see a difference in the HbR concentration between the control and the MS patient in the same region where the topmost influential channels are located, according to Figure 3.

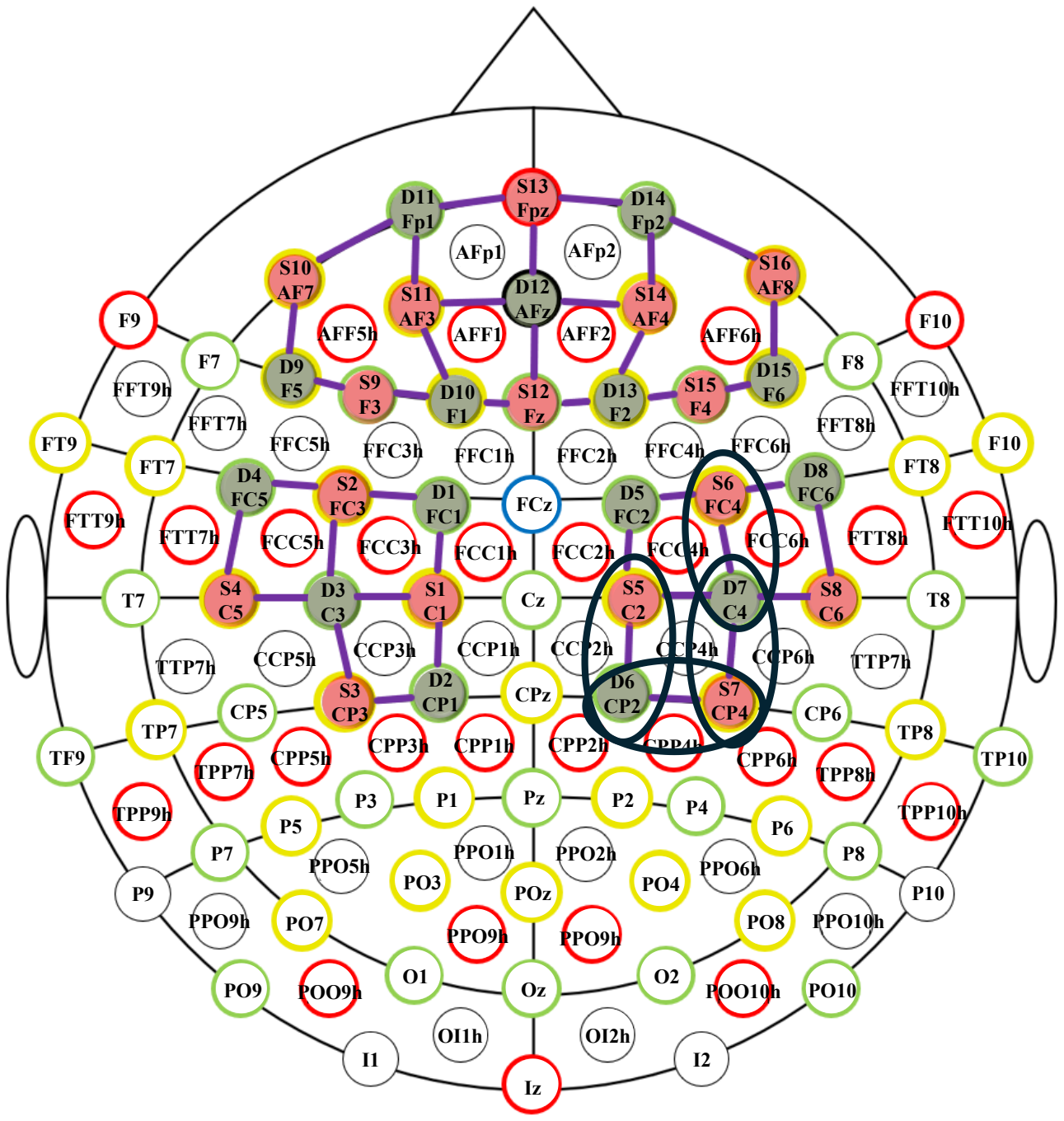

a) Black ovals show the most influential channels for the Single task.

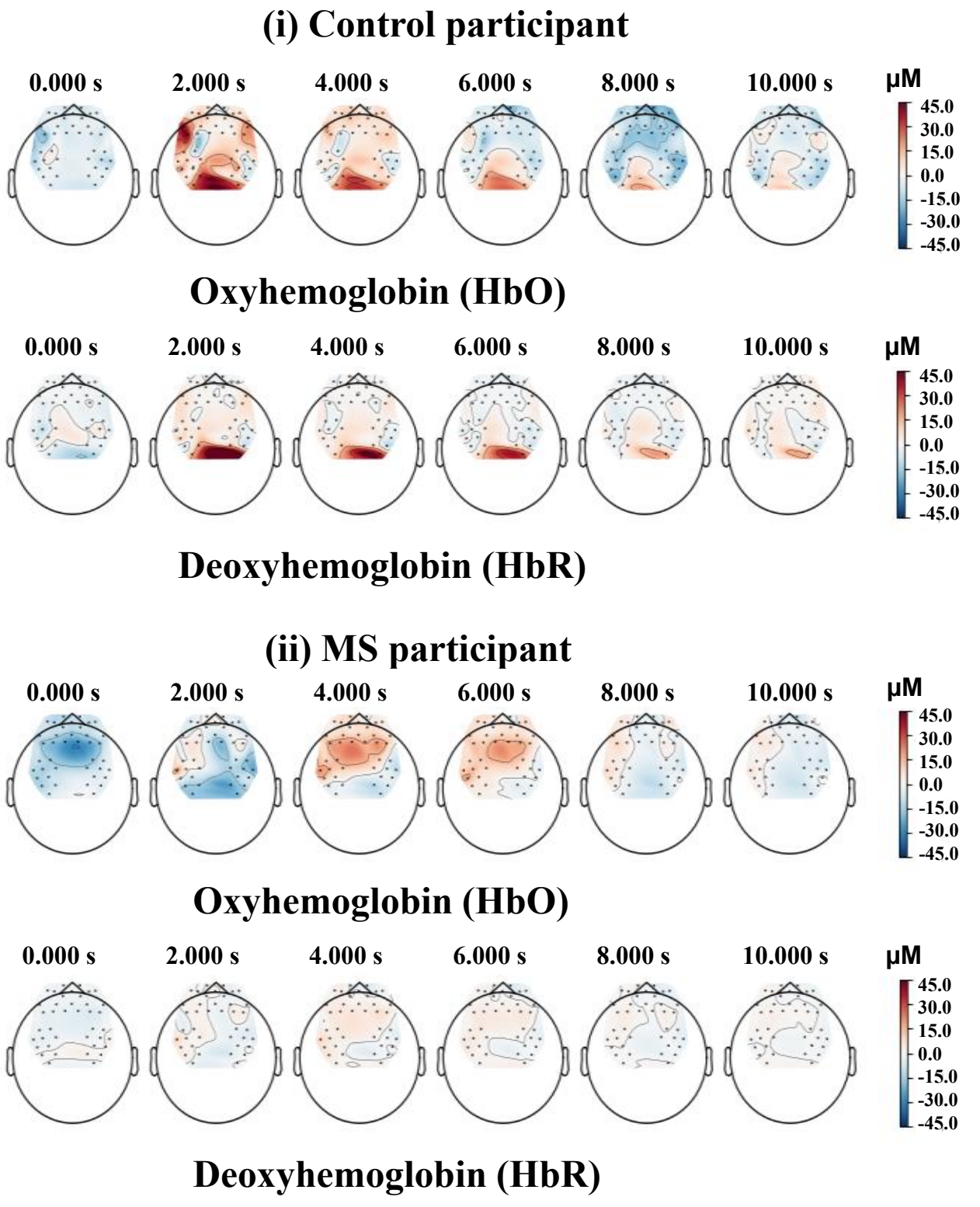

b) Brain activity of Single task for (i) a representative control participant and (ii) a representative MS participant.

Fig. 4. a) Ovals inside the Montage design for the fNIRS cap represent the most influential channels during the single task; b) Topographic Maps of brain activity of one individual control (i) and one individual MS patient (ii) during the single task. The five trials of the single task were averaged into one, and then the plot was generated from 0 to 10 seconds with 2-second intervals, illustrating how brain activity changes over time. The scale on the right indicates the concentration changes in micromoles per liter (μM).

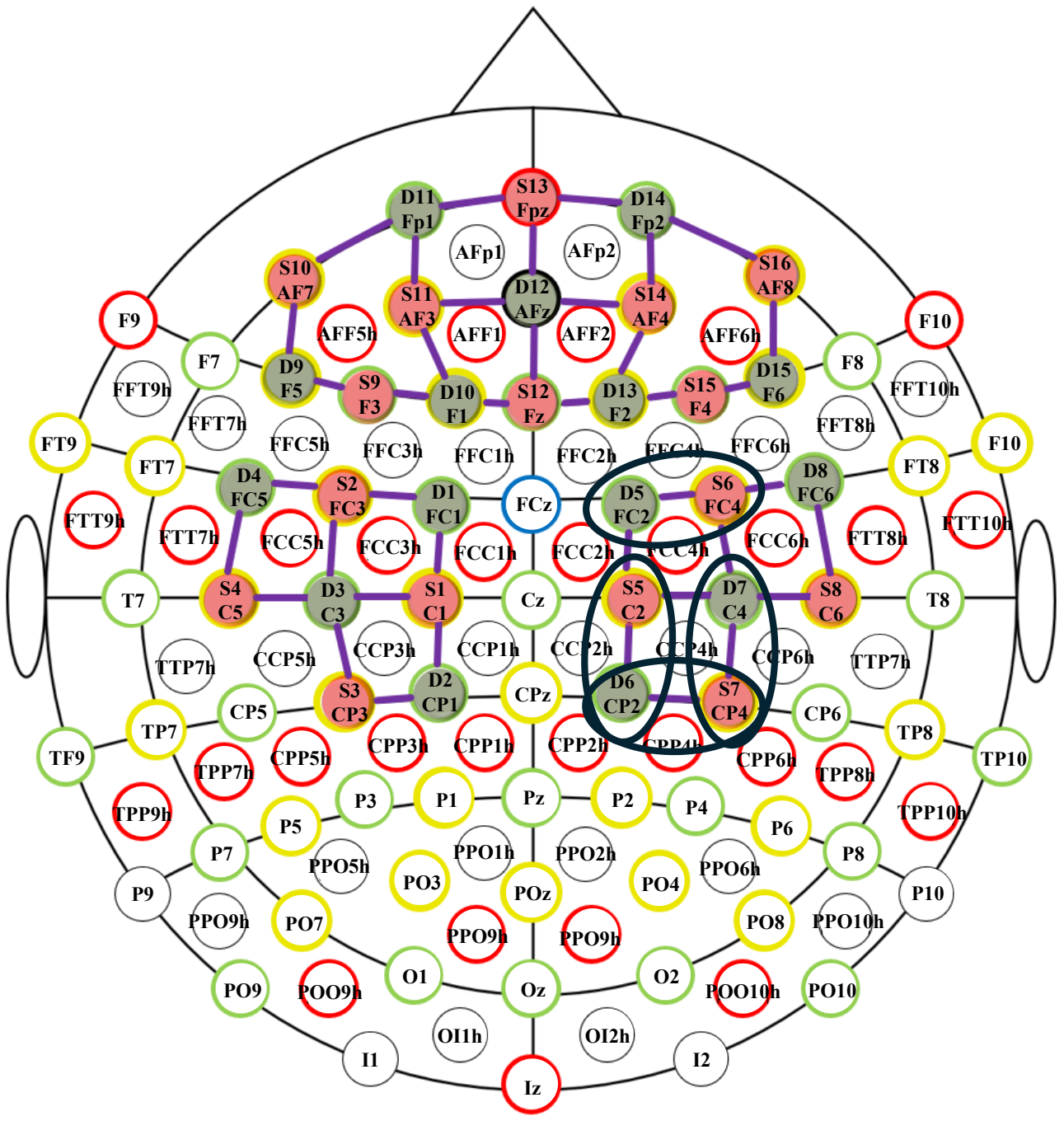

a) Black ovals show the most influential channels for the Dual task.

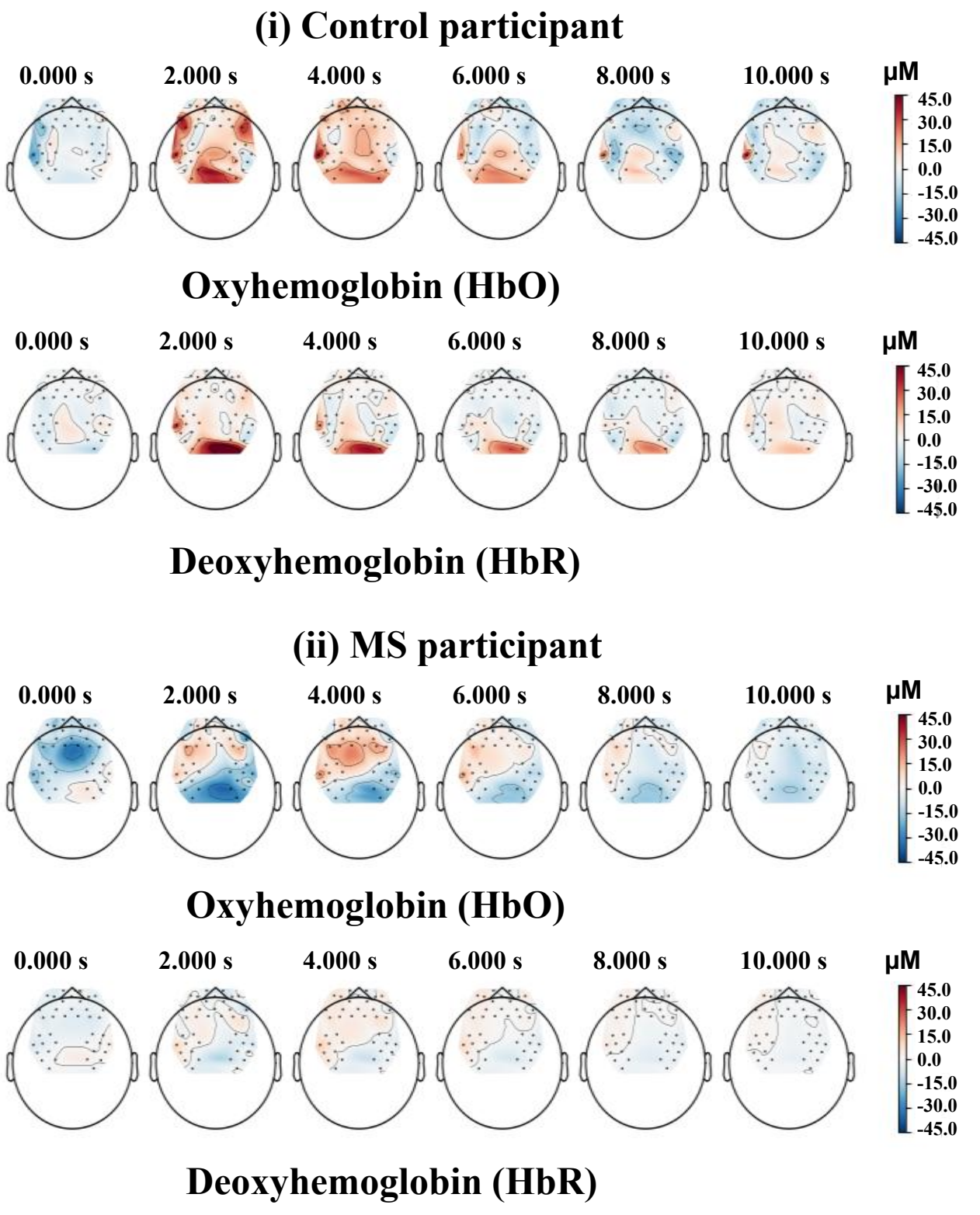

b) Brain activity of Dual task for (i) a representative control participant and (ii) a representative MS participant.

Fig. 5. a) Ovals inside the Montage design for the fNIRS cap represent the most influential channels during the dual task; b) represent the Topographic Map of brain activity over time of one individual control (i) and one individual MS patient (ii) during the dual task, the five trials of the dual task were averaged into one, and then the plot was generated from 0 to 10 seconds with 2 seconds interval illustrating how brain activity changes over time. The scale on the right indicates the concentration changes in micromoles per liter (μM).

In terms of differences in cerebral hemodynamics at the level of each fNIRS channel during the single task (Table IV), only two channels were statistically different between groups among the top four. The MS group had about half the HbR in channel S7-D6 during the single task compared to the controls. According to the 10-20 system of EEG anatomical landmarks [53], [54], this area corresponds to CP2 and CP4, the supramarginal/angular gyri of the right brain hemisphere (responsible for sensory integration) when moving the right hand during the NHPT. At the same time, MS participants showed decreased HbR in channel S5-D6 (ipsilateral to the NHPT movement) compared to controls, an area that corresponds to the precentral gyrus (responsible for voluntary movement). Overall, the results suggest less activation in areas responsible for motor commands and multimodal sensory integration areas in MS participants in the hemisphere ipsilateral to the moving hand.

The results of the dual task (NHPT + holding a ball) showed a similar pattern (Table V), with only two channels being statistically different from the top four, with lesser HbR in channels S7-D6 and S5-D6 in the MS group. Notably, these differences occurred in the hemisphere ipsilateral to the moving hand.

TABLE IV

RESULTS FOR GROUP STATISTICS USING INDEPENDENT SAMPLES T TEST FOR THE TOP FOUR MOST INFLUENTIAL CHANNELS' HEMODYNAMIC RESPONSE MEASURES FOR SINGLE TASK BETWEEN TWO GROUPS. N VALUES FOR CONTROLS = 4336 AND MS = 5135.

| Channels | Group | Mean | Std. Deviation | t | Two-Sided p | Mean Difference |
|---|---|---|---|---|---|---|
| S7-D6 HbR | Control | 1.40E-06 | 8.68E-06 | 5.374 | <.001* | 8.28E-07 |
| | MS | 5.78E-07 | 6.28E-06 | | | |
| S6-D7 HbR | Control | -3.24E-07 | 8.01E-06 | 0.517 | 0.605 | 7.91E-08 |
| | MS | -4.03E-07 | 6.89E-06 | | | |
| S7-D7 HbR | Control | -9.84E-08 | 5.90E-06 | 0.972 | 0.331 | 2.28E-07 |
| | MS | -2.36E-07 | 7.61E-06 | | | |
| S5-D6 HbR | Control | -6.72E-09 | 6.53E-06 | 2.646 | .008* | 3.44E-07 |
| | MS | -3.51E-07 | 6.11E-06 | | | |

TABLE V

RESULTS FOR GROUP STATISTICS USING INDEPENDENT SAMPLES T TEST FOR THE TOP FOUR MOST INFLUENTIAL CHANNELS' HEMODYNAMIC RESPONSE MEASURES FOR DUAL TASK BETWEEN TWO GROUPS. N VALUES FOR CONTROLS = 4336 AND MS = 5135.

| Channels | Group | Mean | Std. Deviation | t | Two-Sided p | Mean Difference |
|---|---|---|---|---|---|---|
| S7-D6 HbR | Control | 1.83E-06 | 8.66E-06 | 5.738 | <.001* | 8.76E-07 |
| | MS | 9.54E-07 | 6.15E-06 | | | |
| S7-D7 HbR | Control | -9.20E-08 | 5.86E-06 | 1.473 | 0.141 | 2.05E-07 |
| | MS | -2.97E-07 | 7.46E-06 | | | |
| S6-D5 HbO | Control | 1.20E-07 | 8.49E-06 | -0.187 | 0.852 | -3.57E-08 |
| | MS | 1.55E-07 | 9.88E-06 | | | |
| S5-D6 HbR | Control | 3.56E-07 | 7.66E-06 | 4.226 | <.001* | 5.80E-07 |
| | MS | -2.24E-07 | 5.68E-06 | | | |

Figures 6 and 7 illustrate the mean hemodynamic responses of the top four channels across the control and the MS groups while performing the single and dual tasks, respectively. The results from Tables IV and V were mostly consistent with Figures 6 and 7, accordingly, whereas, in all four channels, we could see that the MS group had noticeably decreased hemodynamic responses compared to the control group during the task duration of 20s for both tasks.

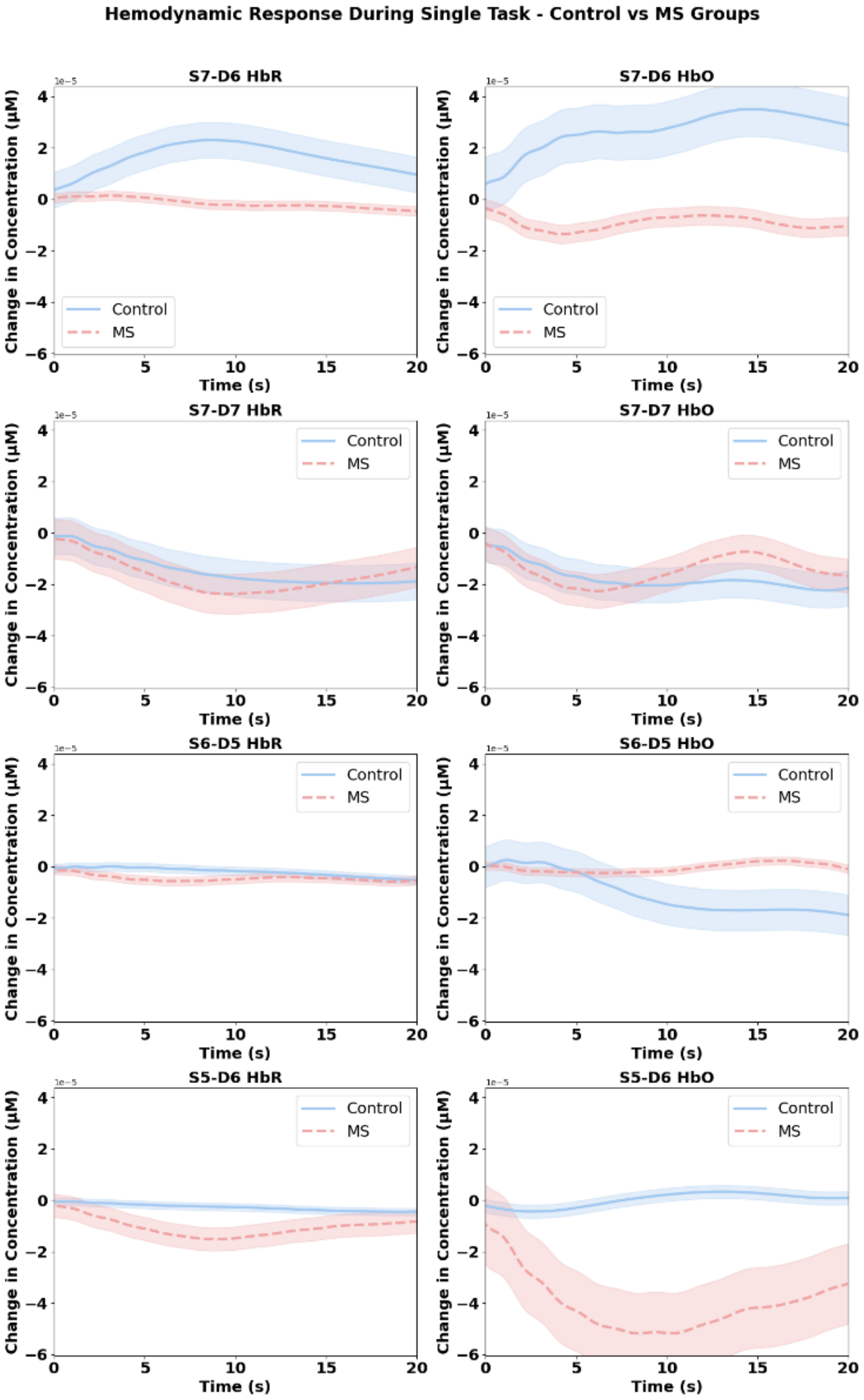

Fig. 6. Hemodynamic responses over time for the top four most influential channels in the control and MS groups during the Single Task, as identified in Table IV. The five trials of the Single Task were averaged into one. Then, the plot was generated for 20 seconds to cover the whole task duration, illustrating how the hemodynamic response changes over time. The units of y-axes are the concentration changes in micromoles per liter (μM). The solid and dashed lines represent the mean hemodynamic response for the control and MS groups; the shaded areas around the lines indicate the standard deviation, reflecting the variability within each group.

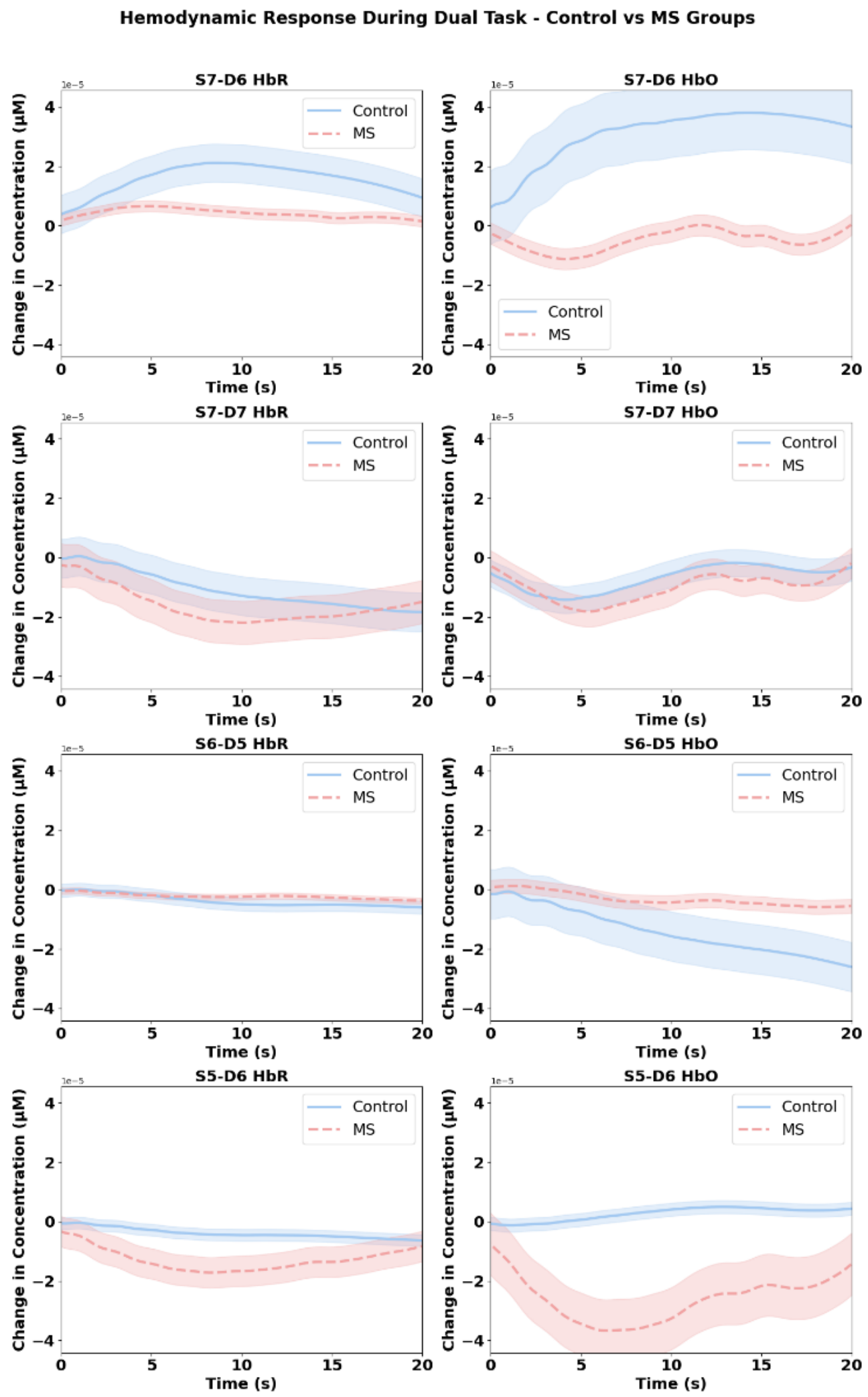

Fig. 7. Visualizes the hemodynamic responses over time for the top four most influential channels in the control and MS groups during the Dual Task, as identified in Table V. The five trials of the Dual Task were averaged into one. Then, the plot was generated for 20 seconds to cover the whole task duration, illustrating how the hemodynamic response changes over time. The units of y-axes are the concentration changes in micromoles per liter (μM). The solid and dashed lines represent the mean hemodynamic response for the control and MS groups; the shaded areas around the lines indicate the standard deviation, reflecting the variability within each group.

Figure 8 demonstrates a comparison of hemoglobin responses (HbO and HbR) between MS patients and the control group during single (Figure 8 (a)) and dual (Figure 8 (b)) tasks in the supramarginal/angular gyri region. Both tasks showed consistent results---the MS group takes, on average, longer than the control group to reach peak deoxygenated hemoglobin levels (HbR). This suggests that MS patients exhibit a slower neurovascular response during complex tasks than controls. We also examined the hemoglobin responses in the precentral gyrus for the controls and MS. However, the control and the MS group did not have any notable differences in reaching the peak HbO or the HbR.

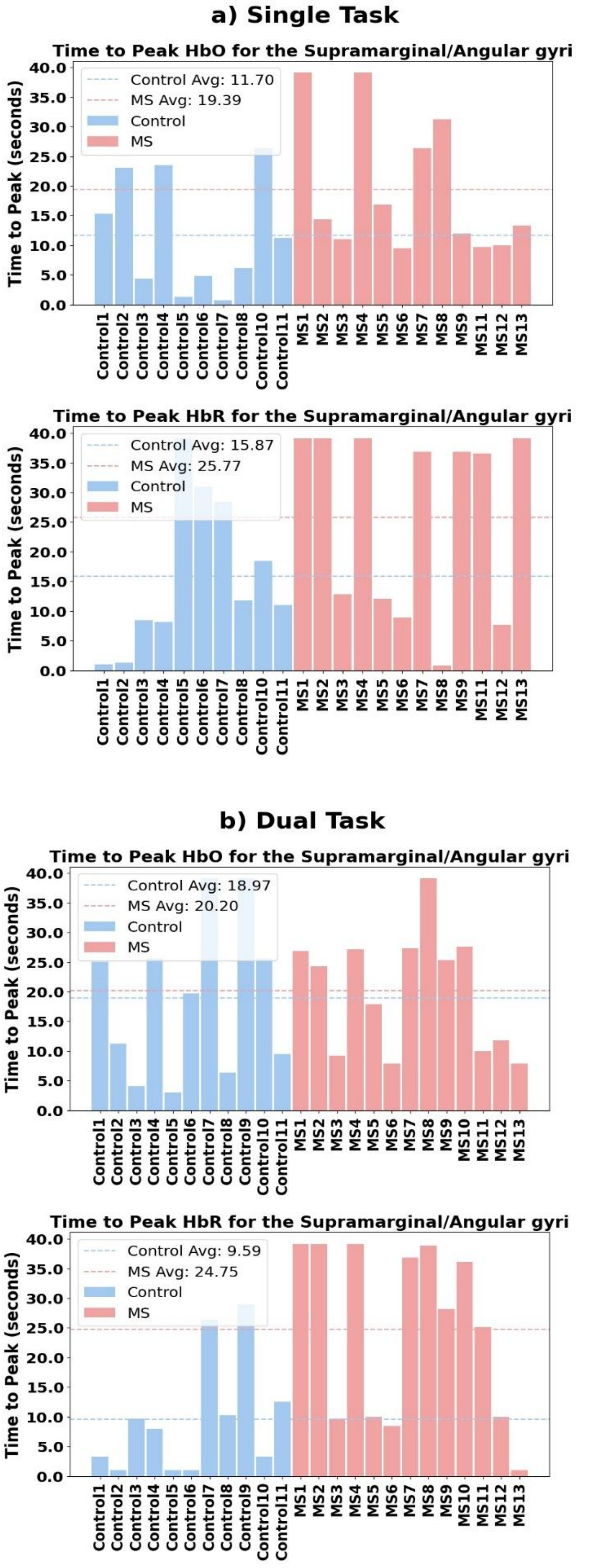

Fig. 8. Comparison of the time to peak hemodynamic responses (HbO and HbR) for the supramarginal/angular gyri between control and MS at the individual level during single and dual tasks, with the supramarginal/angular gyri consisting of channels adjacent to S7,

averaging channels S7-D6 and S7-D7. The top two graphs (a) depict the single task, while the bottom two graphs (b) illustrate the dual task. Each graph displays individual participants' time to peak hemodynamic response, with blue bars representing the control group and red bars indicating the MS group.

## IV. Discussion

The overarching aim of this study was to determine whether explainable AI could be useful in identifying potential brain activity biomarkers that could inform personalized brain stimulation targets for individuals with MS. We employed a robust approach using ML to identify differences in cerebral hemodynamic responses during two types of manual dexterity tasks between MS patients and controls and determine the most important channels (brain regions) that contributed to the ML model. We report three significant findings. First, the ML/AI method effectively determined the key fNIRS channels that contributed to the model's ability to differentiate MS patients from controls. Second, even among individuals with MS without overt hand or cognitive impairment, they differed from controls in cerebral hemodynamic responses in specific fNIRS channels (that corresponded to movement and sensory integration) in the hemisphere ipsilateral to the moving hand. MS patients had lower HbR in the supramarginal/angular gyri and the precentral gyrus (sensory integration and motor regions) of the ipsilateral hemisphere, with suppressed activity and slower neurovascular response. Finally, we observed that HbR response best distinguished the two groups, which was contrary to the conventional evaluation of HbO based on previous recommendations and research [41].

### A. Predicting MS-related changes in brain activity using ML

ML overcomes reductionist data analysis methods such as averaging channels within a region of interest or pre-specifying the specific fNIRS outcome (HbO/HbR, etc.). The data acquired from fNIRS were high-dimensional and in a time sequence. The traditional way of analyzing the data would be to reduce the data into a single data point (mean, peak HbO/HbR, time to peak, etc.) and analyze the data using typical statistical approaches [29]. However, this reductionist approach resulted in losing vital information inside the data. We show that using this method, there appears to be no differences in brain activity between MS and controls (Supplementary Table A). One of the key advantages of the ML approach was that it handled the multivariate and time-series nature of fNIRS data [31], thereby preserving valuable temporal information. Building on ML by utilizing SHAP, we identified two key channels (brain activity biomarkers) from the top four most influential channels (Figure 3) that were statistically significant for the single task (Table IV) and the dual task (Table V). When looking at the brain activity maps of the MS group and the control group for both tasks in Figures 6 and 7, the difference in the hemodynamic response for the top four most influential channels over time between the two groups could be observed clearly. The MS group tends to have a significantly lower activation for all the channels than the control group. Although the traditional approach found no significant difference between the groups (Supplementary Table A), our ML model achieved an accuracy of 75% for the single task and 66.7% for the dual task, which is comparable to those reported in the literature [55]–[57]. As demonstrated in this study, meaningful biomarkers were extracted from the complicated and high-dimensional raw fNIRS data, which are usually difficult for statistical tests or human specialists to analyze directly.

### B. Suppressed ipsilateral activation during dexterous tasks in people with MS

In healthy individuals, movements of the upper limb of the body are controlled mainly through the contralateral hemisphere [62], but complex motor tasks involving the upper limbs elicit bilateral cortical activation [63], primarily involving the sensorimotor cortex, premotor cortex, and supplementary motor areas of the brain. Previous fNIRS studies confirm that tasks completed with the dominant hand primarily use the contralateral sensorimotor cortex while recruiting the bilateral sensorimotor cortex and premotor cortex as tasks become complex [64], [65]. Our findings reveal that individuals with MS have reduced activation in the ipsilateral sensorimotor cortex, despite having comparable performance for NHPT. It is likely that, although our sample of people with MS had no hand impairment (as measured by the NHPT), in fact, their ipsilateral cortical networks were not working in concert with the contralateral hemisphere compared to controls. This suggests that MS-related neurodegeneration may affect brain function before overt behavioral impairments become evident. Such findings support the use of neuroimaging biomarkers to detect early or subtle changes in motor system function that are not captured by conventional performance-based tests. These subtle and covert network inefficiencies may explain why so many people with MS complain of cognitive fatigue when completing everyday tasks.

### C. Slowed neurovascular coupling as measured by time-to-peak

We show evidence that individuals with MS demonstrate prolonged time-to-peak HbO and HbR in the ipsilateral sensorimotor cortex and compared to controls, despite having comparable performance for NHPT. This delay in the hemodynamic response is consistent with evidence of disrupted neurovascular coupling in MS, where the damage to astrocyte-mediated vasodilation pathways causes a potential delay in regional cerebral blood flow regulation [66], [67]. The delayed response to ipsilateral brain activation might reflect exhausted neural compensatory mechanisms due to reduced functional reserve capacity caused by chronic neuroinflammation. ML and Explainable AI were able to identify this ipsilateral hemodynamic latency in the motor command and multimodal sensory integration brain regions as primary features distinguishing between the MS and control group, a phenomenon that has not been explored in prior investigations of subclinical multiple sclerosis. These abnormalities in the subclinical hemodynamics are likely an early biomarker for network inefficiency preceding overt functional decline. Such network inefficiency could contribute to fatigue, a commonly reported MS symptom [68]. This emphasizes the need for more exploration of ipsilateral hemodynamics for individuals with MS.

### D. Discovery of Potential Brain Activity Biomarkers

Brain stimulation, particularly using transcranial magnetic stimulation, is an emerging tool to treat cortical network dysfunction that may underly cognitive fatigue. The challenge in the field is that there are no evidence-based and consistent stimulation targets. For instance, in a pilot trial, brain stimulation over bilateral prefrontal cortices reduced feelings of fatigue in MS [69] , yet in a larger study, similar stimulation over the left dorsolateral prefrontal cortex provided no benefit [16]. It is quite possible that brain stimulation targets vary between individuals, and so novel methods to identify treatment targets/biomarkers, such as those examined in the current study, will help individualize future treatments for MS. The use of explainable AI identified novel targets/biomarkers (supramarginal/angular gyri and precentral gyrus), offering insight into the disrupted neural mechanisms in MS and opportunities for future research.

### E. Usefulness of HbO and HbR as brain activity biomarkers

HbO is a measure of the metabolic demand of activated neurons, while HbR is believed to reflect oxygen utilization. There is substantial variability in fNIRS chromophore reporting with HbO, rather than HbR, the most commonly reported [29]. Our study revealed a compelling finding that HbR provides a clearer distinction of brain activity during tasks than HbO when analyzed using ML techniques. As mentioned above, previous research [58]–[60] utilized a data reductionist approach, specifically Block Averaging [61], where all the trials and rest periods are averaged into one trial and one rest period. Due to this averaging process in the statistical analysis, HbO and HbR might lose their variation in signal and therefore their usefulness as a biomarker.

## V. Limitations and Future Work

Although we specifically targeted people with MS who did not have cognitive or hand impairment (performing better than accepted measurement cut-offs), the MS group was slower on NHPT (mean 21.43s, SD 3.95) compared to controls (mean 18.94s, SD 2.21). They also had slightly lower MoCA score compared to the control group. This suggests that a person with MS may score in the ‘normal’ range on tests that may not be particularly sensitive, and that better biomarkers are required.

We examined brain activation during a dominant hand manual dexterity task; however, we did not use electromyography to examine muscle activation on either hand. It is possible that some participants may have unintentional mirror movement with their non-dominant hand, which we would be unaware of and could have influenced brain activity patterns. It is important to mention that task performance was closely supervised and preceded by a practice session; however, we could not be certain that the non-measured hand remained stationary during the tests.

Finally, ML is very dependent on data. If one does not have sufficient data, then the model will suffer from overfitting [70]. We believe that any ML model would generalize better if trained on a larger dataset. Despite the limitations of ML, it was still a very promising approach over the traditional approaches to the analysis of fNIRS data. Future work should consider whether ML and AI approaches reveal unique brain activation patterns when compared to conventional statistical methods, especially during a wider variety of tasks and disease severity.

## VI. Conclusion

We show that explainable AI, applied to fNIRS data, identified novel brain activity biomarkers that could inform personalized brain stimulation targets for individuals with MS. Using ML, followed by AI, our study identified the differences in cerebral hemodynamic responses during two manual dexterity tasks between stable MS patients having no overt impairments and controls. The most important brain regions contributing to the ML model were the supramarginal/angular gyri and the precentral gyrus (sensory integration and motor regions) of the ipsilateral hemisphere, with suppressed activity and evidence of slowed neurovascular coupling in the MS group. During both tasks, HbR, rather than conventional HbO, the concentration of the channels in the ipsilateral side of the brain distinguished patients from controls. These new brain stimulation targets offer insights into the disrupted neural mechanisms in MS and opportunities for future research and treatment.